\documentclass[a4paper,conference]{IEEEtran}
\ifCLASSINFOpdf
\else
\fi
\usepackage{comment}
\usepackage{multirow}
\usepackage{subfigure}
\usepackage{xcolor}

\newcommand{\Eq}[1]{Eq.~(\ref{eq:#1})}

\usepackage{graphicx}
\usepackage{amsmath}
\usepackage{amssymb}
\usepackage{booktabs}

\newcommand{\ie}{\textit{i}.\textit{e}.}

\usepackage{xspace}

\newcommand{\our}{Zoom-CAM }
\newcommand{\gradcam}{Grad-CAM }
\newcommand{\cam}{CAM }
\newcommand{\etal}{\textit{et al}}

\begin{document}
%
\title{Zoom-CAM: Generating Fine-grained Pixel Annotations from Image Labels}

\author{\IEEEauthorblockN{Xiangwei Shi, Seyran Khademi, Yunqiang Li, Jan van Gemert}
\IEEEauthorblockA{Computer Vision Lab\\
Delft University of Technology, The Netherlands\\
}
}


%


\maketitle

\begin{abstract}
Current weakly supervised object localization and segmentation rely on class-discriminative visualization techniques to generate pseudo-labels for pixel-level training. Such
visualization methods, including class activation mapping (CAM) and Grad-CAM,  use only the deepest, lowest resolution convolutional layer, missing all information in intermediate layers. We propose Zoom-CAM: going beyond the last lowest resolution layer by integrating the importance maps over all activations in intermediate layers.  Zoom-CAM captures fine-grained small-scale objects for various discriminative class instances, which are commonly missed by the baseline visualization methods. We focus on generating pixel-level pseudo-labels from class labels. The quality of our pseudo-labels evaluated on the ImageNet localization task exhibits more than $2.8\%$ improvement on top-1 error. For weakly supervised semantic segmentation our generated pseudo-labels improve a state of the art model by $1.1\%$.

\end{abstract}


%
\IEEEpeerreviewmaketitle

\section{Introduction}
\label{sec:intro}



\begin{figure*}[!t]
 \centering
\includegraphics[width=0.9\textwidth]{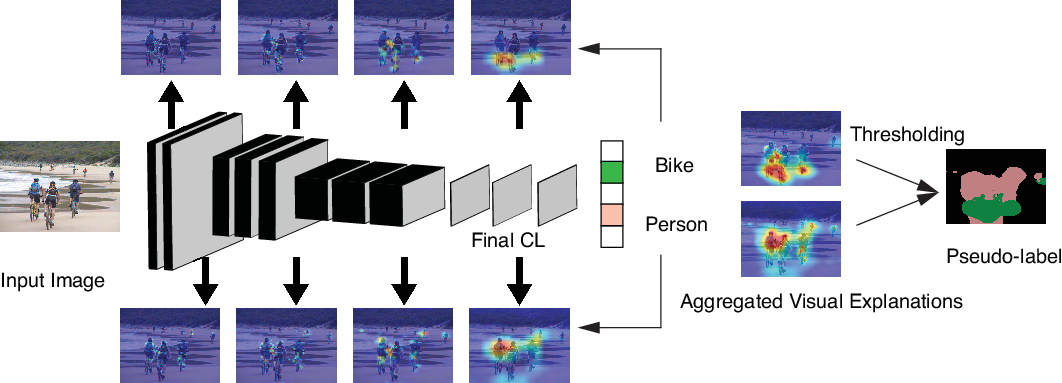}%
\caption{We generate high-precision pixel-level  pseudo-labels for weakly supervised localization and segmentation. We exemplify  pseudo-label generating for an example image with ``bike" and ``person" classes. The pseudo-labels for the intermediate layers are generated by back propagating the gradient of the class score w.r.t. the activations (see Section \ref{sec:met}) demonstrating  localization of the different instances of the class object. Note that the one from only the last convolutional layer over highlights the area around the object and misses the small instances. For clarity, visual explanations are up-sampled to the input size.}
\label{fig:motivational}
\end{figure*}
Visual CNN explanation models allow computer-generated labels (pseudo-labels) to replace laborious human annotations. For example, semantic segmentation~\cite{kolesnikov2016seed,wei2017object,zhang2018adversarial,wei2018revisiting,lee2019ficklenet,laradji2019masks,ahn2019weakly,ahn2018learning,zhou2018weakly,zeng2019joint,wang2019self,wang2019twinsadvnet}  requires expensive pixel-level annotations. Such pixel-level annotations can be generated by CNN visualization methods, with the great advantage of only requiring image-level labels, saving huge annotation costs. Our proposed  method focuses on  generating  fine-grained pseudo-labels from class labels and demonstrated on bounding box labels for object localization and segmentation pixel labels.  

Excellent recent visual explanations such as Score-CAM ~\cite{wang2019score}, Grad-CAM++~\cite{chattopadhay2018grad} and others~\cite{NIPS2017_7062,xu2020attribution} focus on decision faithfulness (causality); yet do not give high-precision localization maps, see Figure \ref{fig:pascal}. This is a problem, as weakly supervised learning methods require fine-grained localization maps to generate  pseudo-labels from class information \cite{SinghL17}. 
Here, we make the observation that current methods use the last convolutional layer (CL) at the lowest resolution. In fact, small objects are eliminated easily after several pooling layers in most CNN models for classification.  Our hypothesis is that by including the visualization maps from intermediate CL, the quality of the pseudo-labels can be improved. 

The last CL offers the most semantically comprehensive spatial information with the smallest dimensions. Moreover, the deconvolution is straightforward when computing the rate of change in the class output with respect to the last CL, as commonly there are few (or none) nonlinear layers in between, to impair the mapping.
Nevertheless, the resolution is severely compromised once the visualization map from the last CL are projected into the input image. This results in coarse visualization with over-highlighted regions in the background or even missing small-scale objects that are completely removed due to several pooling operations. In Grad-Cam \cite{selvaraju2017grad}, there were unsuccessful attempts to go beyond CL as shown in Figure \ref{fig:comparison}. 
 
 In this paper we investigate Zoom-cam: Zoomed-in  pseudo-labels for weakly supervised learned using just class labels. We bridge the visualization between the last CL to the input image by visualizing and integrating not only the last but all the feature maps from intermediate CL. Our focus is to generate fine-grained pseudo-labels that highlight the class objects accurately in the original image, as illustrated in Figure~\ref{fig:motivational}. We have the following contributions:
 \begin{itemize}
    \setlength{\itemsep}{-\itemsep}
    \item \our 
  generates high-resolution visualization maps, capable of identifying several instances of the same class as well as objects with different scales that are often missed by other methods, see Figure~\ref{fig:pascal}.  
 
     \item We introduce an effective gradient back-propagation scheme to obtain weight masks for a linear combination of intermediate feature maps. 
     \item We demonstrate quantitatively that the best explanation belongs to the last CL, yet combining the visualization maps from intermediate layers reduces the noise corresponding to the locality of the gradient flows. 
\item On the ImageNet localization task we show 2.8\% and 3.7\% improvement on top-1 and top-5 errors, respectively, when the objects are localized using \our visual explanations compared to \gradcam.   

 \item   
 By plugging our method in a state of the art weakly supervised model~\cite{ahn2019weakly} it improves by $1.1\%$  where our mIoU for visual explanations outperforms  \cite{chattopadhay2018grad,wang2019score,selvaraju2017grad}.
 \end{itemize}
 
 In the rest of this paper, we refer to the heatmaps  that are resized to the input image (See Figure~\ref{fig:motivational}) as \textit{visual explanations}. The importance weight matrix for aggregated feature maps in CL are called \textit{weigh masks}.   Note that  \textit{activation units} and \textit{activation neurons} are used interchangeably.







%


\section{Related work}
\label{sec:rw}
 \textbf{Automatic Pseudo-labels Generation.} 
Our work focuses on generating high-precision visual explanations from class label information to be used for weakly supervised object localization and segmentation models. Earlier practices to retrieve localization information from class labels utilize the intermediate feature maps. Oquab \etal~\cite{oquab2014learning} show the object localization ability of image classification CNNs by transferring mid-level image representations using a global max pooling layer. 
Current weakly supervised object segmentation and localization methods~\cite{kolesnikov2016seed,wei2017object,zhang2018adversarial,wei2018revisiting,lee2019ficklenet,laradji2019masks,ahn2019weakly,ahn2018learning,zhou2018weakly,zeng2019joint,wang2019self,wang2019twinsadvnet,gudiBMC17objectExtentPool,Yang2019CombinationalCA} take advantage of visualization techniques such as CAM and Grad-CAM to automatically generate pseudo labels for training purposes.
  Therefore, the quality of the pseudo labels generated by visualization methods majorly affects the segmentation performance. In this context, a precise and complete  visual explanations is of great importance.  
  Wei \etal~\cite{wei2018revisiting} aggregated multiple CAM visualizations generated by dilated CNNs with different rates to obtain more accurate regions in visual maps. Lee \etal~\cite{lee2019ficklenet} randomly selected hidden units from CNNs to generate multiple localization maps that highlight different parts of the class discriminative objects.   
  We evaluate the quality of the generated visualizations by Zoom-Cam, using a weakly supervised segmentation CNN model proposed in~\cite{ahn2019weakly} as a measure for the precision of visual explanations. the quality of the generated visual explanations is improves in \cite{SinghL17}  by hiding different parts of the input while generating the visual explanations using CAM. In \cite{SinghL17} the final pseudo-labels for object and action localization are generated by aggregating several explanations, in the cost of larger computations. 
   
   To the best of our knowledge, all weakly supervised segmentation and localization methods rely on  visual explanation techniques to generate the pseudo-labels. We are the first to target completeness in visual explanations for weakly supervised tasks.

\textbf{Visualizing CNNs.} Many  explainability work~\cite{wang2019score,chattopadhay2018grad,zeiler2014visualizing,petsiuk2018rise,viering2019manipulate} focuses on the faithfulness of the visual explanations to the decision made by a CNN. We noticed that even though these methods are serving their purpose to spot the causal features, they lack the precision and completeness required for accurate pixel-level localization.  
We discuss related literature on visualizations of CNNs in the following. 
 Inspired by  \cite{Du2019TechniquesFI,Zhang2018VisualIF}, we describe three categories:
1) Deconvolution methods 2) Blind methods and 3) Representation methods. 

\emph{Deconvolution methods} ~\cite{springenberg2014striving,zeiler2014visualizing,simonyan2013deep} aim at mapping the maximum of activation units in CLs, back to the original images. Zeiler and Fergus~\cite{zeiler2014visualizing}, in a pioneering work, attempted to interpret the CNN filters by deconvolution  to identify what input patterns lead to the maximum activation. Springenberg \etal~\cite{springenberg2014striving} proposed guided backpropagation to visualize the feature maps in networks without max-pooling layers (fully convolutional).  Even though easily interpretable, Deconvolution methods generate visualizations within several forward and backward passes and thus computationally expensive. The visual explanations generated by Deconvolution  are not fine-grained, due to the non-linear nature of CNN models that makes the inverse mapping impossible.  

\emph{Blind methods} follow the black box approach, where the system between input and output is assumed inaccessible. Blind methods ~\cite{zhou2014object,dabkowski2017real,petsiuk2018rise,wagner2019interpretable} generate visual explanations by perturbing (setting pixel intensities to zero, blurring the region or by adding noise) the input in pixel-space to measure the variation in model prediction score. The input regions that increase the classification score (no spatial cues) are reflected in highly activated feature maps with spatial information.  ~\cite{zhou2014object} covered small-region inputs with patches to identify the highly activated units in the receptive fields. \cite{petsiuk2018rise} use random binary masks on the entire image. 
Having infinite options for variations of input space, motivated blind methods to integrate perturbation into loss function~\cite{fong2017interpretable,dabkowski2017real}.
 \cite{fong2017interpretable} backpropagated the gradients of feature maps to learn a perturbation mask for the input space. Blind methods are known to be reliable and faithful to the underlying model as they capture the CNN response w.r.t. the global changes imposed on the input. Nevertheless,  testing different range of variations in the input space impose a great computational burden for generating visual explanations.   



\clearpage

 \emph{Representation  methods} ~\cite{Du2018TowardsEO,10.1145/2939672.2939778,selvaraju2017grad,chattopadhay2018grad,zhou2016learning,wang2019score} generate visual explanations 
 based on gradients and/or weights under the assumption that these are accessible.
 This is in contrast to the blind approach.
 Commonly, in representation method the gradient flow is used as a local measure that captures the variation of the output w.r.t. the features. 
  A popular technique is the class activation mapping (CAM) proposed by Zhou \etal~\cite{zhou2016learning}, which generates visual explanations as a linear weighted combination of activation maps of the last convolutional layer. To do so, the model architecture has to change by replacing the fully connected layers with a global average pooling (AP) layer and subsequent retraining. Selvaraju \etal~\cite{selvaraju2017grad} proposed gradient-weighted class activation mapping (Grad-CAM), by using the average gradients of target class w.r.t. the last convolutional feature maps to compute the CAM weights without re-training the CNN model. Grad-CAM++~\cite{chattopadhay2018grad}, the variant of~\cite{selvaraju2017grad}, aims at generating more accurate localization maps by individually weighting the activation units in the last feature map, instead of
  the global average pooling in \cite{zhou2016learning,selvaraju2017grad}. Score-CAM~\cite{wang2019score} is an input-perturbation-based variant of CAM that measures the class posteriors directly yet relying on the activation units in the ultimate layer.  
  Our \our is the first to exploit intermediate convolutional feature maps for generating pseudo-labels used for weakly supervised learning.

\section{Methodology}
\label{sec:met}

\begin{figure*}[t!]
    \centering
    \subfigure[\scriptsize{Input image}]{
    \begin{minipage}[]{0.13\textwidth}
    \includegraphics[width=1\textwidth]{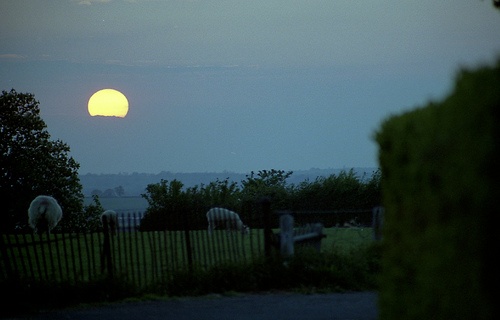}
    \includegraphics[width=1\textwidth]{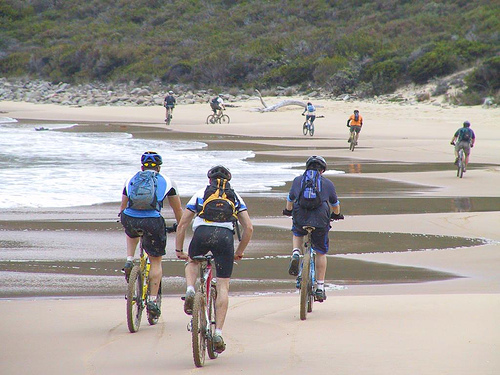}
    \includegraphics[width=1\textwidth]{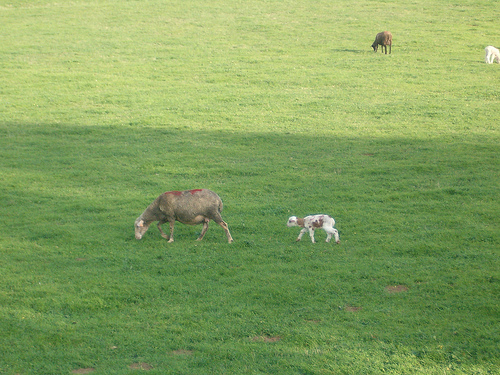}
    \includegraphics[width=1\textwidth]{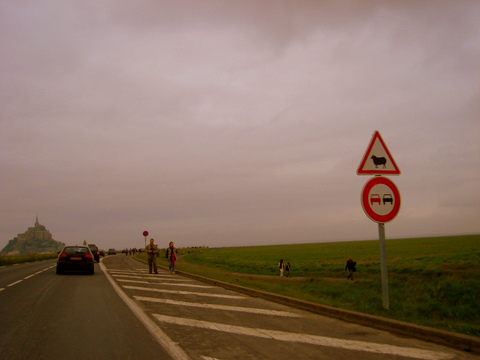}
    \includegraphics[width=1\textwidth]{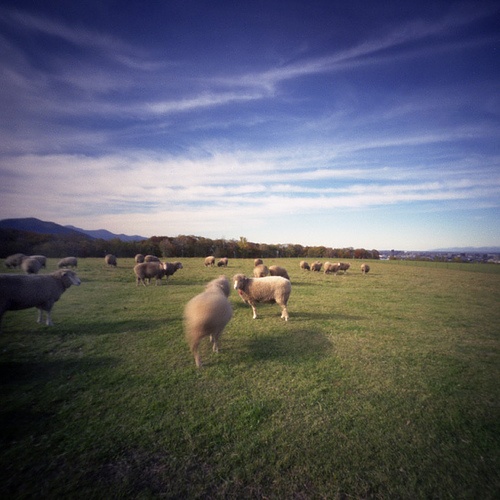}
    \includegraphics[width=1\textwidth]{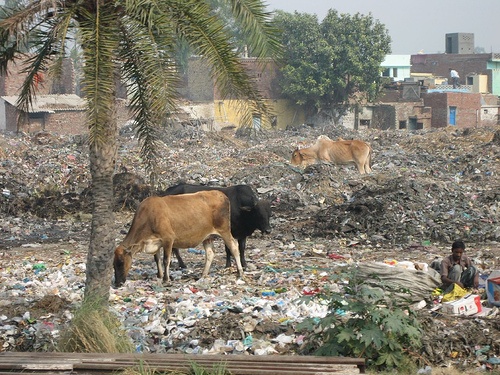}
    \includegraphics[width=1\textwidth]{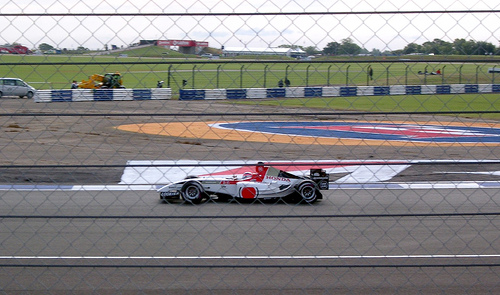}
    \includegraphics[width=1\textwidth]{images/2011_001020.jpg}
    \end{minipage}
    }
    \subfigure[\scriptsize{Ground truth}]{
    \begin{minipage}[]{0.13\textwidth}
    \includegraphics[width=1\textwidth]{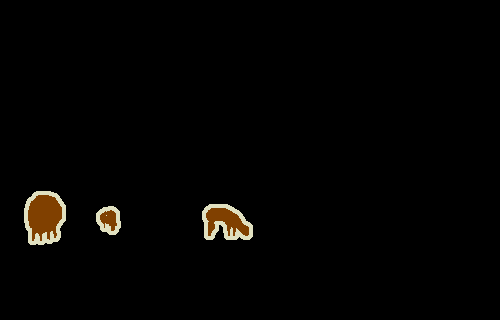}
    \includegraphics[width=1\textwidth]{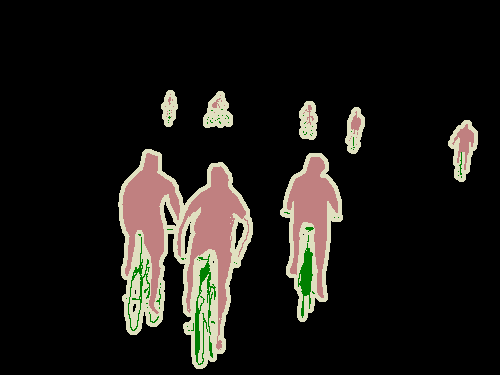}
    \includegraphics[width=1\textwidth]{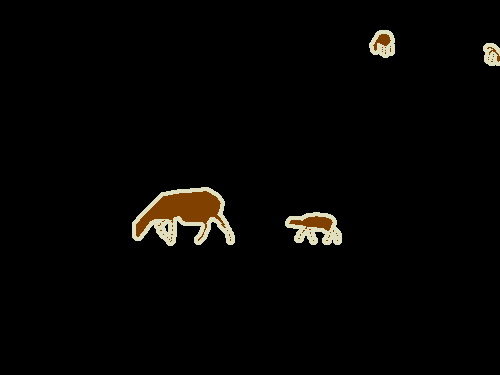}
    \includegraphics[width=1\textwidth]{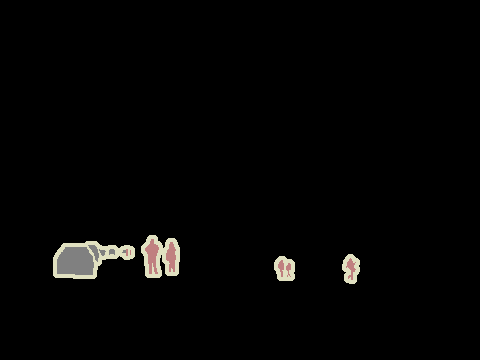}
    \includegraphics[width=1\textwidth]{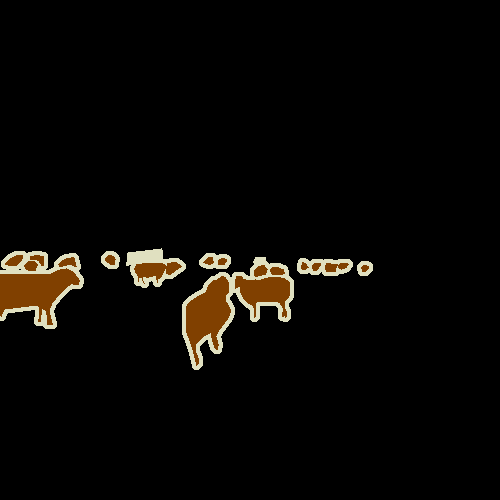}
    \includegraphics[width=1\textwidth]{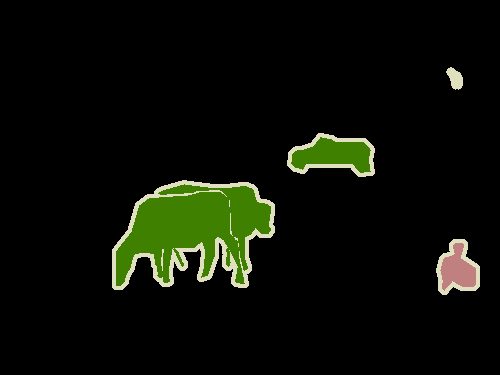}
    \includegraphics[width=1\textwidth]{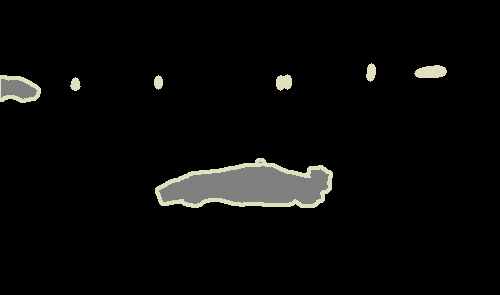}
    \includegraphics[width=1\textwidth]{images/2011_001020.png}
    \end{minipage}
    }
    \subfigure[\scriptsize{Zoom-CAM}]{
    \begin{minipage}[]{0.13\textwidth}
    \includegraphics[width=1\textwidth]{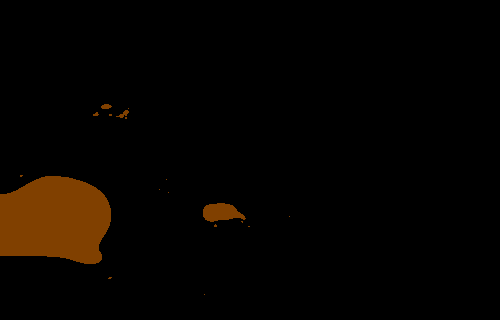}
    \includegraphics[width=1\textwidth]{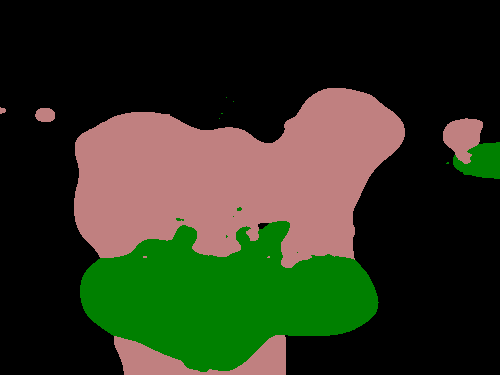}
    \includegraphics[width=1\textwidth]{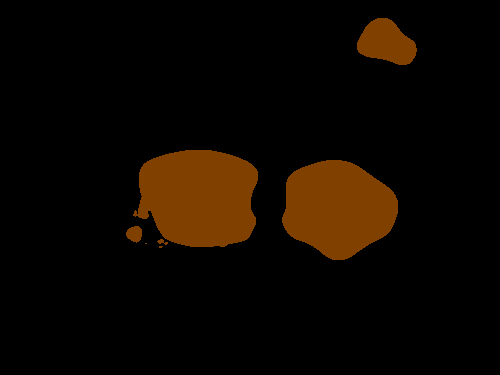}
    \includegraphics[width=1\textwidth]{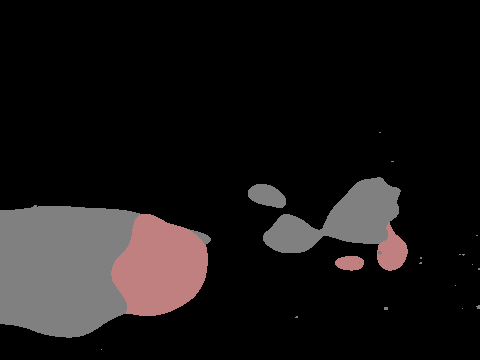}
    \includegraphics[width=1\textwidth]{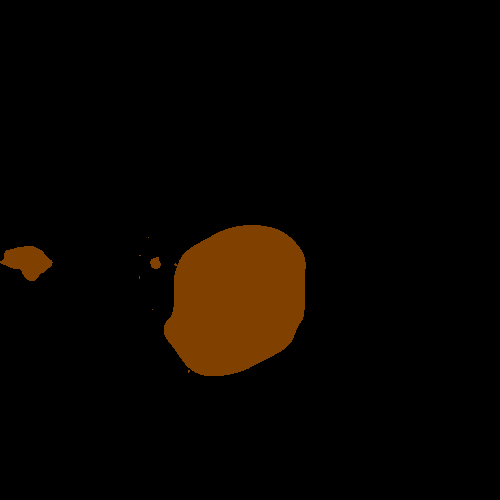}
    \includegraphics[width=1\textwidth]{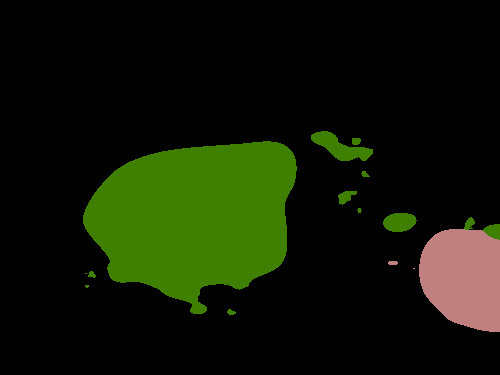}
    \includegraphics[width=1\textwidth]{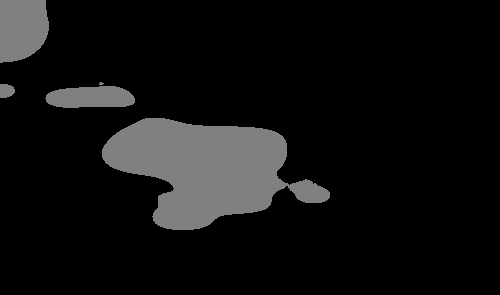}
    \includegraphics[width=1\textwidth]{images/2011_001020_ours.png}
    \end{minipage}
    }
    \subfigure[\scriptsize{Grad-CAM}]{
    \begin{minipage}[]{0.13\textwidth}
    \includegraphics[width=1\textwidth]{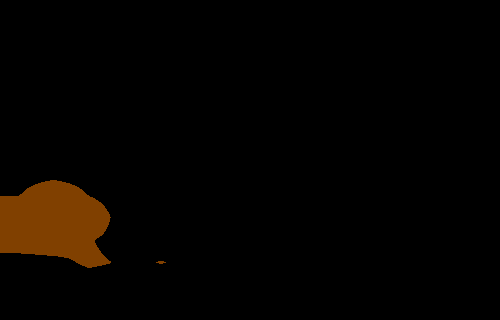}
    \includegraphics[width=1\textwidth]{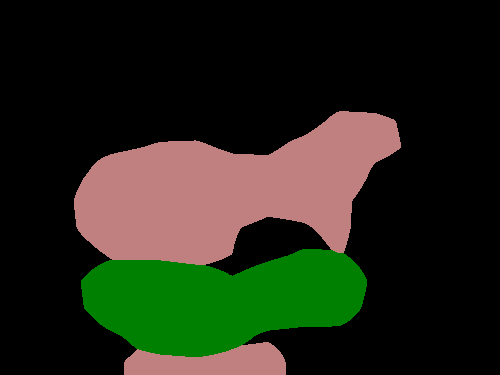}
    \includegraphics[width=1\textwidth]{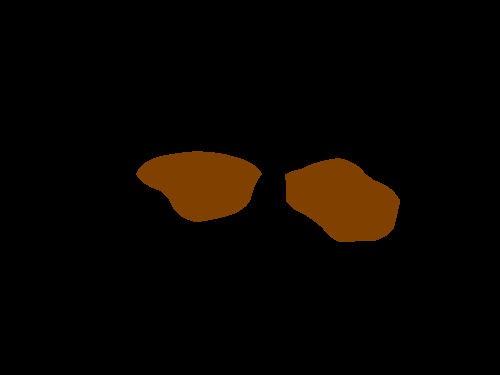}
    \includegraphics[width=1\textwidth]{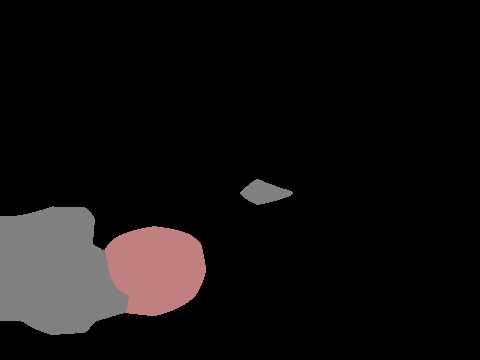}
    \includegraphics[width=1\textwidth]{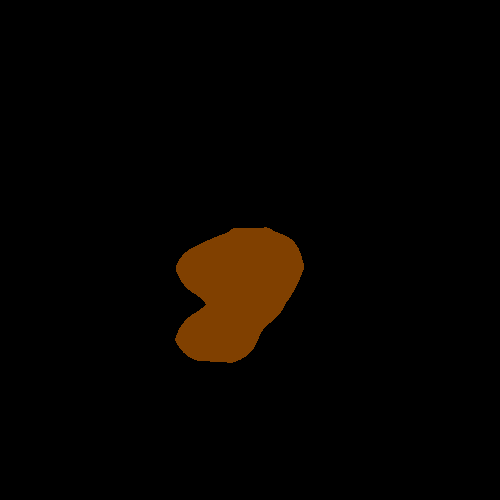}
    \includegraphics[width=1\textwidth]{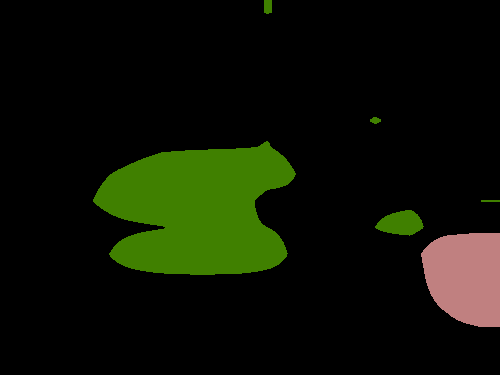}
    \includegraphics[width=1\textwidth]{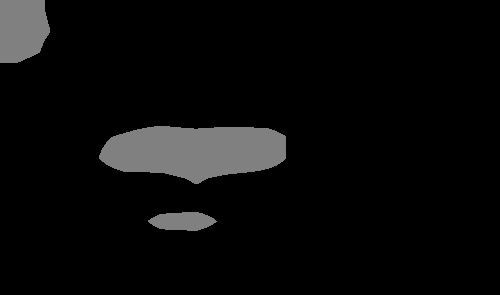}
    \includegraphics[width=1\textwidth]{images/2011_001020_Grad-cam.png}
    \end{minipage}
    }
    \subfigure[\scriptsize{Score-CAM}]{
    \begin{minipage}[]{0.13\textwidth}
    \includegraphics[width=1\textwidth]{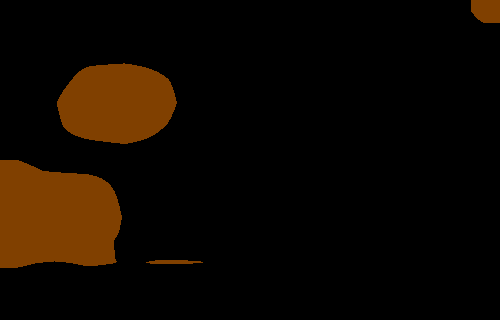}
    \includegraphics[width=1\textwidth]{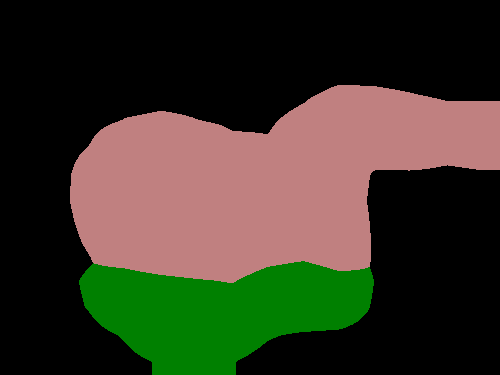}
    \includegraphics[width=1\textwidth]{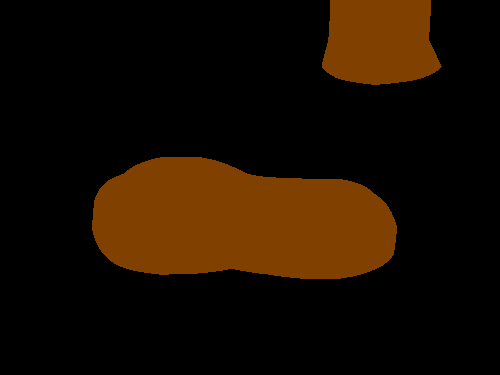}
    \includegraphics[width=1\textwidth]{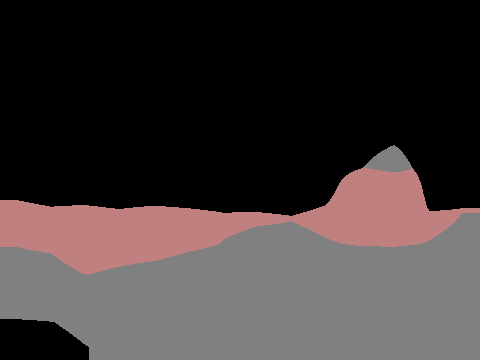}
    \includegraphics[width=1\textwidth]{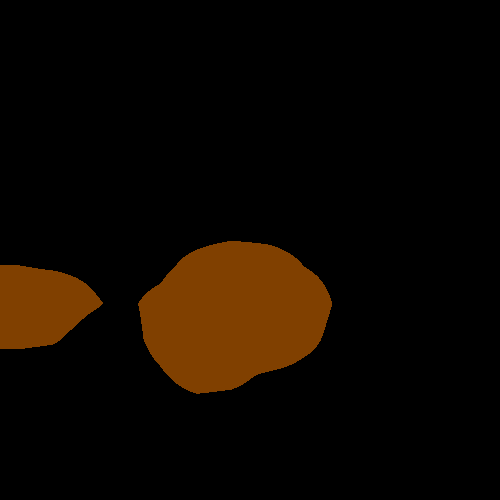}
    \includegraphics[width=1\textwidth]{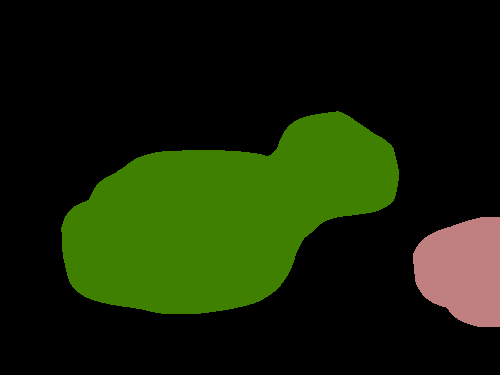}
    \includegraphics[width=1\textwidth]{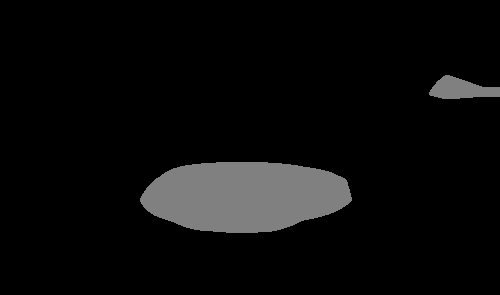}
    \includegraphics[width=1\textwidth]{images/2011_001020_scorecam.png}
    \end{minipage}
    }
    \subfigure[\scriptsize{Grad-CAM++}]{
    \begin{minipage}[]{0.13\textwidth}
    \includegraphics[width=1\textwidth]{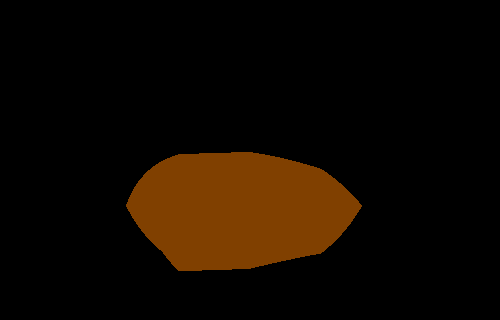}
    \includegraphics[width=1\textwidth]{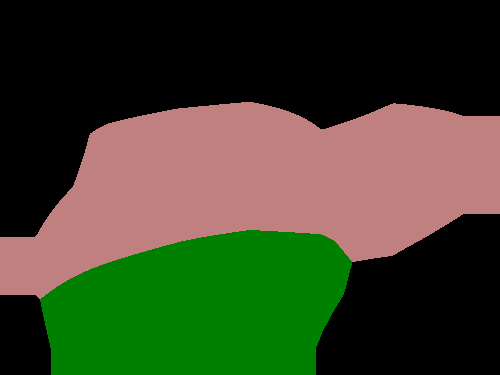}
    \includegraphics[width=1\textwidth]{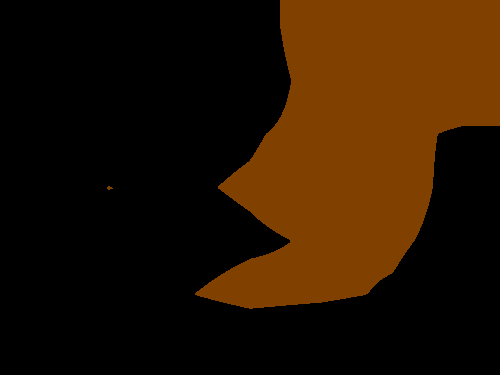}
    \includegraphics[width=1\textwidth]{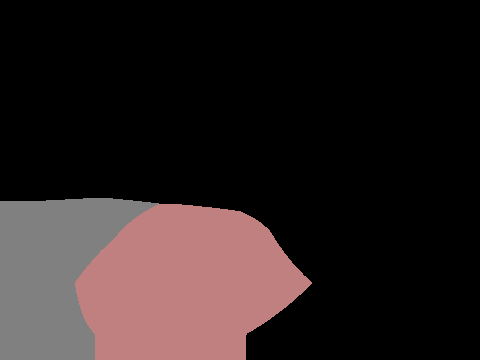}
    \includegraphics[width=1\textwidth]{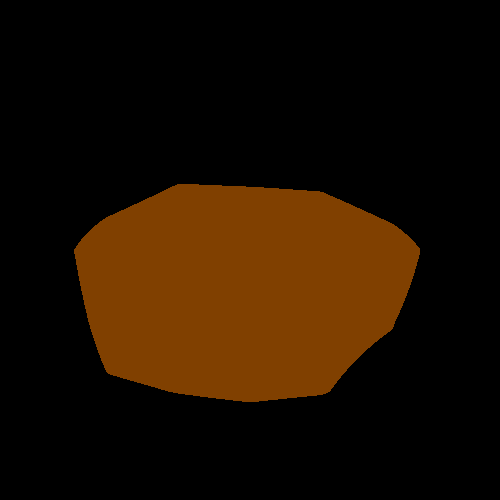}
    \includegraphics[width=1\textwidth]{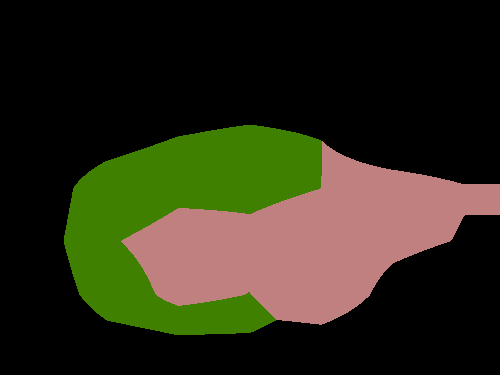}
    \includegraphics[width=1\textwidth]{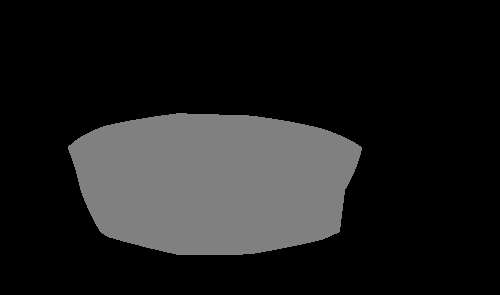}
    \includegraphics[width=1\textwidth]{images/2011_001020_gradcam++.png}
    \end{minipage}
    }
    \caption{Examples of pseudo-labels generated by Zoom-CAM visual explanations (ours), Grad-CAM \cite{selvaraju2017grad}, Score-CAM \cite{wang2019score} and Grad-CAM++\cite{chattopadhay2018grad}. Zoom-CAM aggregates visual maps through all intermediate layers, which captures objects with different scales and several instances of the same class. The over-highlighted regions relate to false positive. Zoom-CAM can generate fine-grained pseudo-labels by increasing the true positive and reducing the false positive.}
    \label{fig:pascal}
\end{figure*}

\begin{figure*}[t]
\centering
    \begin{minipage}[]{0.95\textwidth}
       \subfigure[]{
    \includegraphics[width=0.45\textwidth]{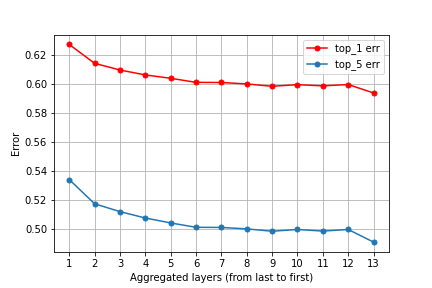}
    }
        \subfigure[]{
    \includegraphics[width=0.45\textwidth]{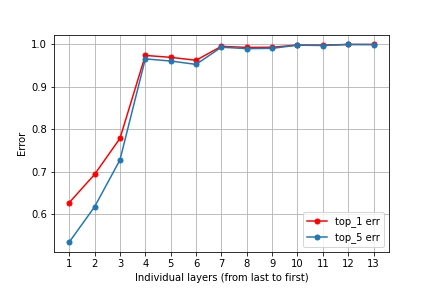}
    
    }
    \end{minipage}
    \caption{Top-1 and top-5 localization error rates on ILSVRC2012 \textit{val} dataset for ablation study. 
    {(a)} Aggregating intermediate feature maps can consistently improve the weakly supervised object localization ability, especially when the last two layers are integrated. 
    {(b)} The localization error rates of Zoom-CAM maps using feature maps from a single intermediate layer. The feature maps of the last layer contributes the most to the performance of object localization.
    } 
    \label{fig:imagenet_ablation}
\end{figure*}


Our proposed method is inspired by Grad-CAM~\cite{selvaraju2017grad} and its variants, and we use the backward gradients from the class score (before softmax) to weight the neurons in the feature maps. In addition, by extending the gradient flow beyond the LC, \our enables the visualization of any intermediate layers in CNN. This extension is not straightforward and is explained in the following

\subsection{Revisiting CAM and Grad-CAM}
Suppose $A_k(i,j)$ is the $i,j$-th activation in the $k$-th feature map of the last convolutional layer. Following the definition in  
\cam and \gradcam, the final score for the class $c$, before softmax, is the weighted sum of the average pooled activation neurons in the last convolutional layer: 
\begin{equation}
\label{eq:CAM}
S^c := \sum_k\alpha_k^c\sum_{i,j}A_k(i ,j),
\end{equation}
where $\alpha_k^c$ are the \cam weights once the network is retrained by replacing the fully connected layers with a global AP layer. \gradcam shows that $\alpha_k^c$ can be replaced with the average gradient of the class score w.r.t. the neurons in $A_k$. Thus, retraining is not required and the gradients can be obtained by single backward pass operation. 
Accordingly, the visualization map for \gradcam is given by: 
\begin{equation}
\label{eq:gradcam_map}
L_{i,j}^c := \text{ReLU}\big(\sum_k\,\,\frac{1}{Z}\sum_{i,j} \frac{\partial{S^c}}{\partial{A_k(i,j)}}\, A_k(i,j)\,\big),
\end{equation}
where $Z$ is the number of activation units in the feature maps of the last convolutional layer and $S^c$ is the score for class $c$. The ReLU function in \Eq{gradcam_map} guarantees that only neurons with positive contribution to the gradient of the class score are considered. Note that by defining $L_{i,j}^c$ from \Eq{CAM},  the sum of the elements in $L_{i,j}^c$ is guaranteed to be equal to the class score $S^c$. 
\gradcam averages the gradients in the feature maps to get $\alpha_k^c$ as weights for linear combination of feature maps.
Instead, we use the back propagated gradients as a {\textit{weight mask}} (matrix) applied to the feature maps. Careful reformulation of the problem is essential for extracting the weigh masks as explained in the following.   
\subsection{Visualizing Intermediate Layers}
Here we go beyond Grad-Cam and describe how Zoom-Cam visualizes the intermediate layers. Suppose $B_p(m,n)$ is the $m,n$-th activation in the $p$-th feature map of the penultimate convolutional layer. Based on the common operational units in a forward pass of a CNN model, $B_p$ is passed through a non-linear function $f$ such as ReLU, sigmoid, tanh, etc. Then $f(B_p)$ for all $p={1,2,...,P}$ in the penultimate convolutional layer is convolved with the $k$-th filter and summed over $p$ to generate the last convolutional feature map, \ie, $A_k$. Given that convolution is a linear operation, one can write the sum over the activation units in $A_k$ as a weighted sum of $f(B_p(m,n))$:  
\begin{equation}
\label{eq:sum_Ak}
\sum_{i,j}A_k(i ,j) = \sum_{m,n} W_k(m,n)\sum_{p}f(B_p(m ,n)).
\end{equation}
The elements in matrix $W_k$ are the sum over subset of filter weights
for $k$-th kernel. In turn, the filter weights in each subset is a function of $m,n$-th position in $B_p$, the size of the kernel, the stride and the padding of the convolution. 

Substituting \Eq{sum_Ak} in \Eq{CAM} yields the following 
\begin{equation}
\label{eq:class_score}
S^c := \sum_k\alpha_k^c\,\sum_{m,n} W_k(m,n)\sum_{p}f(B_p(m ,n)).
\end{equation}
Similar to \Eq{gradcam_map}, the visualization map of the penultimate convolutions is defined by removing the summation over $m,n$ in \Eq{class_score}, resulting in 
\begin{equation}
\label{eq:zoomcam_map}
L_{m,n}^c := \sum_k\,\alpha_k^c\, W_k(m,n)\sum_{p}f(B_p(m ,n)).
\end{equation}

Note that the nonlinear function $f$ including the ReLU function, batch-norm, or pooling layer, is fixed in the backward pass as we are not in the training process. Thus, $f(B_p)$ can be replaced with a Hadamard product of matrix $N_p \in \mathbb{R}^{m,n}$, representing the nonlinear operation, and $B_p \in \mathbb{R}^{m,n}$. Let us define $F^c \in \mathbb{R}^{m,n}$ as the matrix representation of the penultimate convolutional layer after non-linearity in the backward pass  
\begin{equation}
\label{eq:Fc}
F^c(m,n):= \sum_{p}N_p(m,n)(B_p(m ,n)).
\end{equation}
From the chain rule, one can write the gradient of the final score for class $c$ w.r.t. the represented penultimate convolutional layer  as 
\begin{equation}
\label{eq:gradiant_Fc}
\frac{\partial{S^c}}{\partial{F^c(m,n)}}= \frac{\partial{S^c}}{\partial{\sum_{i,j}A_k(i ,j)}}\times \frac{\partial{\sum_{i,j}A_k(i ,j)}}{\partial{F^c(m,n)}}.
\end{equation}
The right side of \Eq{gradiant_Fc} is carefully decomposed to two terms. The first term is the scaled weights of \gradcam layer and the second term can be derived from \Eq{sum_Ak}. Particularly, for a ReLU activation function, the elements of matrix $N_p$ in \Eq{Fc} are zeros and ones, thus, \Eq{gradiant_Fc} boils down to  
\begin{equation}
\label{eq:gradiant_Fc_cal}
\frac{1}{Z}\,\frac{\partial{S^c}}{\partial{\sum_p B_p(m',n')}}=   \alpha_k^c \, W_k(m',n').
\end{equation}
where $m',n'$ indicates the positive elements that are passed by ReLU function.  
Comparing \Eq{zoomcam_map} and \Eq{gradiant_Fc_cal}, reveals that the weights for visualization map of the penultimate convolutional layer are the gradients of the final score w.r.t. the features. We refer to the $W_k(m',n')$ as weight masks that are applied (point-wise multiplication) to the feature maps. This is in contrast to the \gradcam approach which uses the scalar weighting of the feature maps. 

The final visual explanation for \our,
is calculated by considering only the positive values in $L_{m,n}^c$ because we are only interested in the activation neurons whose intensity should be increased in order to increase the class score.
\begin{equation}
\label{eq:final_map}
L_{m,n}^c := \text{ReLU} \big(\frac{1}{Z}\,\sum_k \sum_{p}\, \frac{\partial{S^c}}{\partial{\sum_p B_p(m' ,n')}} B_p(m' ,n')\big).
\end{equation}
\Eq{final_map} can be extended for any intermediate CNN layer,
by replacing $B_p$ by that layer. 
In \gradcam, a single backward pass up to the last CL, is performed to calculate the gradients. In practice, $W_k(m,n)$ is computed in the backward pass from the AP layer of \gradcam to the target layer. 

\begin{figure*}[htb]
    \centering
    \subfigure[Grad-CAM]{
    \begin{minipage}[]{1\textwidth}
    \includegraphics[width=1\textwidth]{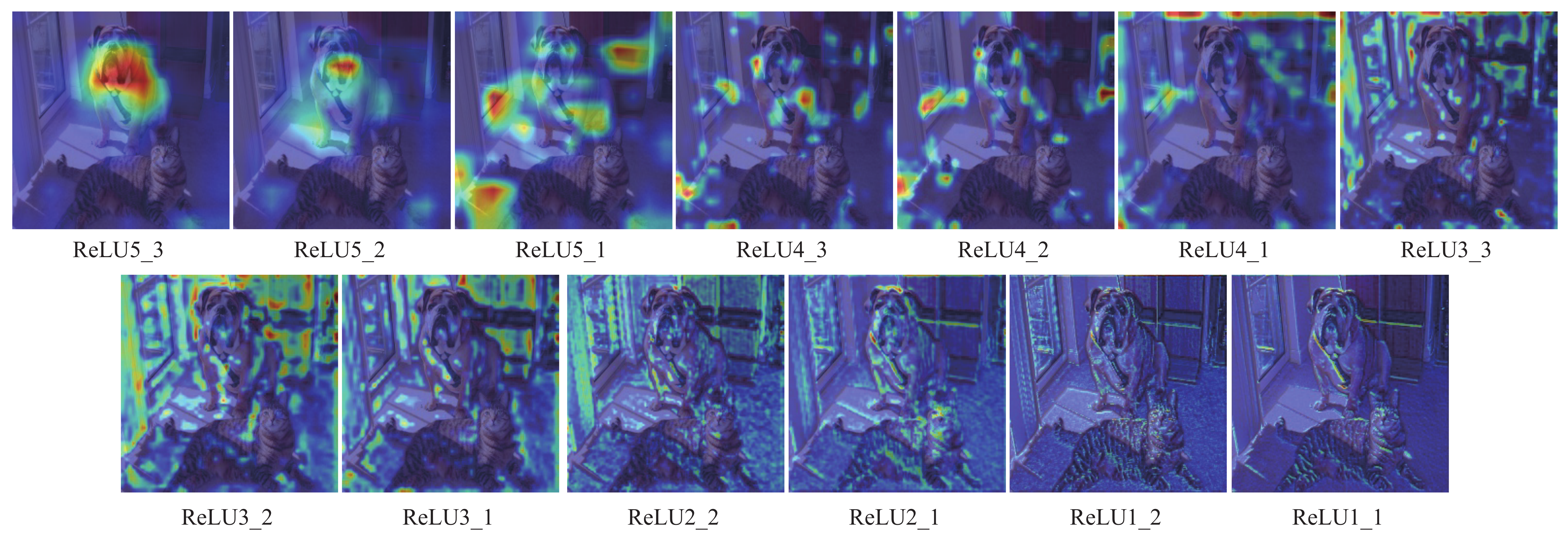}
    \end{minipage}
    }
    \subfigure[Zoom-CAM]{
    \begin{minipage}[]{1\textwidth}
    \includegraphics[width=1\textwidth]{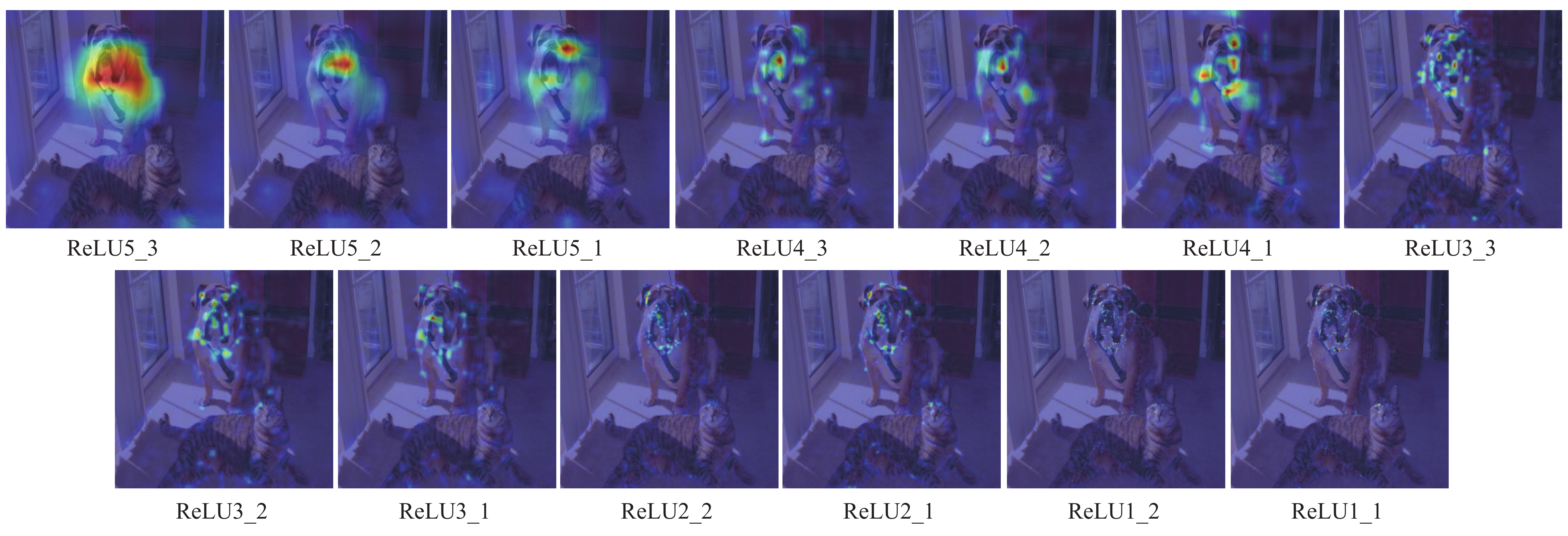}
    \end{minipage}
    }
    \caption{Comparison of visual explanation of single intermediate convolutional layer generated by Grad-CAM and Zoom-CAM. The images for each method represents using the feature maps from the last convolutional layer to the first one. For the same original image (both dog and cat), these saliency maps are generated w.r.t the 'bull mastiff' label. The basic model is pre-trained VGG16 model from PyTorch. Zoom-CAM is using \Eq{gradiant_Fc_cal} to calculate the different weights for different elements of intermediate feature maps, while Grad-CAM takes the average of \Eq{gradiant_Fc_cal} for all elements of an intermediate feature maps.}
    \label{fig:comparison}
\end{figure*}

A \gradcam \cite{selvaraju2017grad} extension visualizes intermediate convolutional layers by average pooling the elements of an intermediate feature maps. In contrast, Zoom-CAM is using \Eq{gradiant_Fc_cal} to calculate individual weights for different elements of intermediate feature maps. The visual explanations for different intermediate layers produced by \gradcam and \our are presented in Figure \ref{fig:comparison}.

\subsection{Aggregation of Localization Maps}
After generating intermediate layer visualization maps of Zoom-CAM via \Eq{final_map}, we need to aggregate these maps and up-sample them to the input image resolution.

Given two Zoom-CAM visualization maps for different intermediate layers, $L^c_{i,j}$ and $L^c_{m,n}$, where $i \leq m$ and $j \leq n$, the first step is the normalization. visualization maps are normalized such that the values over single localization map range from 0 to 1.
Next, we up-sample $L^c_{i,j}$ (smaller feature map) to the size of $L^c_{m,n}$ through bilinear interpolation. Finally, the aggregated visualization maps $L^c$ will be obtained by:
\begin{equation}
\label{eq:aggregation}
\hat{L}^c_{m,n} = \max\{ \text{\textbf{\textit{N}}}(L^c_{m,n}), \text{\textbf{\textit{U}}}(\text{\textbf{\textit{N}}}(L^c_{i,j}))\},
\end{equation}
where $\textit{\textbf{U($\cdot$)}}$ and $\textit{\textbf{N($\cdot$)}}$ denotes the up-sampling and normalization operations, respectively.

Taking the maximum in \Eq{aggregation} is a simple operation that will preserve the importance of visualization maps, that is reflected by the normalized values. Taking the average is another option but we observed the smoothing of the weights across the layers, which is not desirable for generating crisp visualization maps.

\vspace{-0.3cm}

\begin{table}[t]
\centering
\small
\caption{Classification and localization error rates (\%) on ILSVRC2012 val dataset for pre-trained VGG16 from PyTorch. Zoom-CAM by aggregating visualization maps from intermediate layers achieves better performance on object localization than Grad-CAM. The thresholds for Zoom-CAM and Grad-CAM maps are 25\% and 15\% of the maximum value. Zoom-Cam achieves lower error.}
 \setlength{\tabcolsep}{3.5mm}{
\begin{tabular}{ccccc}
\toprule
          & \multicolumn{2}{c}{Classification error}  \ \ & \multicolumn{2}{c}{Localization error} \\
         & Top-1            & Top-5          \ \  & Top-1           & Top-5           \\ \cmidrule(lr){2-3} \cmidrule(lr){4-5}
Zoom-CAM  & 31.87            & 11.54            \ \ & \textbf{59.11}           & \textbf{48.64}           \\
Grad-CAM & 31.87            & 11.54           \ \  & 61.95           & 52.35           \\ \bottomrule
\end{tabular}
	\vspace{0.05in}

\label{table:imagenet}
}
\end{table}

\begin{table*}[t]
\centering
\caption{Comparison of quality of pseudo-segmentation-labels of PASCAL VOC 2012 \textit{val} set measured in IoU (\%).
The base model is a fine-tuned ResNet50, trained on image class labels. The accuracy of Zoom-CAM pseudo-labels compares favorably to others. 
}
\setlength{\tabcolsep}{0.9mm}{
\begin{tabular}{ccccccccccccccccccccccc}
\toprule
\multirow{2}{*}{Method} & \multicolumn{21}{c}{IoU} & \multirow{2}{*}{mIoU} \\ 
 & \rotatebox{90}{backgr} & \rotatebox{90}{plane} & \rotatebox{90}{bike} & \rotatebox{90}{bird} & \rotatebox{90}{boat} & \rotatebox{90}{bottle} & \rotatebox{90}{bus} & \rotatebox{90}{car} & \rotatebox{90}{dog} & \rotatebox{90}{chair} & \rotatebox{90}{cow} & \rotatebox{90}{dtable} &\rotatebox{90}{cat} & \rotatebox{90}{horse} & \rotatebox{90}{motor} & \rotatebox{90}{person} & \rotatebox{90}{plant} & \rotatebox{90}{sheep} & \rotatebox{90}{sofa} & \rotatebox{90}{train} & \rotatebox{90}{tv}\\ \cmidrule(lr){2-22}
Grad-CAM++ & 64.7 & 27.8 & 17.8 & 25.0 & 23.8 & 31.6 & 47.2 & 38.8 & 46.6 & 18.4 & 42.1 & 32.5 & 40.8 & 40.0 & 41.6 & 32.2 & 26.8 & 39.6 & 33.3 & 42.1 & 32.9 & 35.5\\
Grad-CAM & 66.5 & 29.7 & 18.3 & 25.5 & 19.3 & 33.6 & 51.0 & 42.4 & 49.0 & 19.2 & 41.2 & 36.7 & 41.6 & 40.5 & 43.6 & 41.9 & 28.9 & 39.8 & 34.2 & 39.3 & 36.5 & 37.1 \\
Score-CAM & 68.1 & \textbf{31.8} & 19.1 & \textbf{29.7} & \textbf{29.3} & 30.9 & 50.3 & \textbf{45.3} & 47.9 & 19.8 & 41.8 & 32.3 & \textbf{44.7} & 42.0 & \textbf{47.2} & 35.4 & 27.9 & 42.8 & 36.6 & \textbf{47.1} & 31.8 & 38.2\\
Zoom-CAM & \textbf{68.9} & 31.0 & \textbf{19.7} & 26.9 & 20.6 & \textbf{34.5} & 50.3 & 42.3 & \textbf{50.1} & \textbf{20.4} & \textbf{45.6} & 35.3 & 43.2 & \textbf{43.8} & 46.0 & 42.0 & \textbf{31.1} & \textbf{45.0} & \textbf{38.3} & 40.1 & \textbf{38.6} & \textbf{38.8} \\
\bottomrule
\end{tabular}
}
\vspace{0.05in}
\label{tab:miou}
\end{table*}

\begin{table*}[t]
\centering
\small
\parbox{.39\linewidth}{{
\centering
\caption{Quality of pseudo semantic segmentation labels in mIoU, evaluated on the augmented PASCAL VOC 2012 \textit{train} set.
    }
 \setlength{\tabcolsep}{4mm}{
    \begin{tabular}{cc}
\toprule
       Method  & mIoU \\ \midrule
      CAM  &  48.3~\cite{ahn2019weakly} \\
      Zoom-CAM & \textbf{49.0} \\ \bottomrule
    \end{tabular}}
    
    \label{tab:weakly_pseudo}
}}
\hfill
\parbox{.58\linewidth}{{
\centering
\caption{Semantic segmentation performance in mIoU evaluated on the PASCAL VOC 2012 \textit{val} set. The performance of WSSS using pseudo-labels generated by Zoom-CAM is better than the one by CAM.}
 \setlength{\tabcolsep}{4.8mm}{
  \begin{tabular}{cc}
\toprule
        Method &\textit{val}  \\ \midrule
        IRNet(ResNet50)-CAM & 63.5 \\ 
        IRNet(ResNet50)-Zoom-CAM & \textbf{64.6} \\ \bottomrule
    \end{tabular}}
    
    \label{tab:wsss}
}}
\end{table*}

\section{Experiments}
We evaluate the quality of the generated visual explanations by Zoom-Cam. We mostly follow the 
validation framework from \gradcam and \cam by evaluating on weakly supervised object localization and segmentation tasks, on ImageNet and PASCAL VOC datasets, respectively.
Moreover, sample visualization are reported in the supplementary material for qualitative inspection of the results. 

\label{sec:exp}
\subsection{Weakly Supervised Object Localization}
We evaluate  weakly supervised object localization using the visual explanations generated by Zoom-CAM. We use a pre-trained VGG16  as a baseline on the ILSVRC2012~\cite{russakovsky2015imagenet} val dataset. 
We resize images to $224\times224\times3$ and color normalize the mean and the standard deviation. 
We generate Zoom-CAM saliency map in addition to the class prediction. 
The pixels with higher value than $25\%$ of the max intensity are preserved, which constructs several connected regions. We keep the largest connected component and draw a bounding box around it. This bounding box reveals the location of the classified object. We follow the evaluation metrics of ILSVRC2012 object localization task and report the top-1 and top-5 classification and localization error in Table~\ref{table:imagenet}. For localization score, the prediction counts when the classification prediction matches the ground truth image label and the predicted bounding box has over 50\% overlap with the ground truth bounding box.
The results on the ImageNet localization task exhibits around $2.84\%$ improvement on top-1 error after aggregating all intermediate layers. For both \our and \gradcam we use the same CNN model for classification and therefore the classification scores are the same. 
%
%


%
\subsubsection{Ablation Studies} Zoom-CAM aggregates feature maps of all 13 intermediate layers in VGG16. We conduct ablation experiments by aggregating different numbers of intermediate layers including only a single intermediate layer. This is to quantify the contribution of each layer to the accuracy of generated visualizations by \our in terms of weakly supervised object localization.  Fig~\ref{fig:imagenet_ablation} (a) shows that aggregating intermediate layers consistently improves the performance of weakly supervised object localization, especially when the last two layers are integrated. Fig~\ref{fig:imagenet_ablation} (b) shows the top-1 and top-5 localization errors for Zoom-CAM using the feature maps of only single intermediate layer. As expected the last feature map contributes the most to the performance in object localization. 
\subsection{Weakly Supervised Semantic Segmentation}
We evaluate the visualization maps generated by \our on weakly supervised semantic segmentation (WSSS) task on PASCAL VOC 2012~\cite{everingham2010pascal} dataset. Although the dataset contains semantic and instance segmentation labels, we only take advantage of image-level class labels. The training set for  semantic segmentation is augmented by~\cite{hariharan2011semantic}, which contains 10,582 images. The original \textit{val} set with 1,449 images are used for validation.

The task of weakly supervised segmentation leverages the image-level class information to segment objects, including semantic and instance segmentation. Recent works on weakly supervised segmentation use CAM or Grad-CAM to generate pseudo-segmentation-labels for training purposes. Therefore, weakly supervised segmentation models are sensitive to the quality of generated pseudo-labels by the visualization techniques. We first compare the quality of pseudo-labels obtained by Zoom-CAM and 
other visualization methods.
\subsubsection{Quality of Pseudo-labels}
To evaluate the quality of pseudo-labels, we generate saliency maps for each image in the \textit{val} set of PASCAL VOC 2012 
via Zoom-CAM. We take pre-trained ResNet50 on ImageNet~\cite{russakovsky2015imagenet} as the base model and fine-tune it on the classification set of PASCAL VOC 2012.
The mean average precision (mAP) is 94.1\% for the classification task evaluated on \textit{val} set of PASCAL VOC 2012 classification task.
Similarly, we take a threshold, 25\% of the max intensity for Zoom-CAM, on the saliency maps and search the largest connected component. Because images have multiple labels, we threshold and fuse
the saliency maps by comparing saliency values of multiple labels pixel-wise as the final pseudo-labels. 

Table~\ref{tab:miou}
shows the results for pseudo-segmentation-labels using mean Intersection of Union (mIoU) as evaluation metric. For a single class, the quality of pseudo-labels is evaluated by Intersection of Union (IoU).
We can see that adding intermediate featuremaps by Zoom-Cam compares favorably to others.
%

%
Fig~\ref{fig:pascal} shows sample visual explanations w.r.t. the semantic object segmentation ground truth. These examples confirm that Zoom-CAM  captures objects with different sizes, which are commonly lost in the last low-resolution convolutional layer. Interestingly, \gradcam outperforms its recent variants such as Score-CAM and Grad-CAM++ once inspected visually in
PASCAL VOC 2012 dataset. This is consistent with the results in Table~\ref{tab:miou} where pseudo-labels generated by Grad-CAM++ achieve lowest mIoU. 
\subsubsection{Training the Weakly Supervised Semantic Segmentation (WSSS) Baseline with \our Pseudo-labels}
We re-train the s.o.t.a. weakly supervised semantic segmentation model \cite{ahn2019weakly} with generated psudo-labels 
by \our.
We show in Table~\ref{tab:miou} that Zoom-CAM generates pseudo-labels with higher quality so we expect to see improvement in WSSS baseline when trained by \our pseudo-labels. \cite{ahn2019weakly} trains ResNet50 from scratch for the classification task of PASCAL VOC 2012. Then they use CAM to generate pseudo-labels for the training of their segmentation model.  
We replace CAM pseudo-labels with Zoom-CAM pseudo-labels and re-train the WSSS CNN model referred to as IRNet in~\cite{ahn2019weakly}. Table~\ref{tab:weakly_pseudo} shows the quality of pseudo-semantic-segmentation labels in mIoU, evaluated on the PASCAL VOC 2012 segmentation \textit{train} set. The quality of pseudo-labels generated by Zoom-CAM is better than the ones generated by CAM reported in~\cite{ahn2019weakly}, therefore we expect better performance of IRNet on segmentation task once trained with \our pseudo-labels. Finally, Table~\ref{tab:wsss} shows the performance of IRNet using pseudo-labels generated by Zoom-CAM and CAM, which confirms our speculation. This is the ultimate experiment to quantify the effect of more precise visualization maps in a down-stream task such as WSSS. We observed that the mIoU evaluated on PASCAL VOC 2012 \textit{val} set for the re-trained model by \our pseudo-labels improved by 
1.1\%.

\vspace{-0.3cm}

~\\

\section{Conclusion}
\label{sec:con}
We presented \our  to generate high-quality pseudo-labels by integrating visual maps over all intermediate layers in classification CNNs. \our is  a generalization of \gradcam but differently we use weight masks to linearly combine the feature maps at any intermediate CL. The results verify our hypothesis that intermediate layers offer more accurate localization of the object, in CNNs. The computation time of Zoom-CAM visualization is mostly dominated by the time of back-propagation, which is the same as Grad-CAM. We would like to evaluate the faithfulness of our generated visual explanations to the model prediction as well.



\bibliographystyle{IEEEtran}
\bibliography{bare_conf.bib}


\end{document}